# Hierarchical Exploration for Accelerating Contextual Bandits

**Yisong Yue** YISONGYUE@CMU.EDU
iLab, H. John Heinz III College, Carnegie Mellon University, Pittsburgh, PA 15213, USA

**Sue Ann Hong** SAHONG@CS.CMU.EDU
**Carlos Guestrin** GUESTRIN@CS.CMU.EDU
School of Computer Science, Carnegie Mellon University, Pittsburgh, PA 15213, USA

## Abstract

Contextual bandit learning is an increasingly popular approach to optimizing recommender systems via user feedback, but can be slow to converge in practice due to the need for exploring a large feature space. In this paper, we propose a coarse-to-fine hierarchical approach for encoding prior knowledge that drastically reduces the amount of exploration required. Intuitively, user preferences can be reasonably embedded in a coarse low-dimensional feature space that can be explored efficiently, requiring exploration in the high-dimensional space only as necessary. We introduce a bandit algorithm that explores within this coarse-to-fine spectrum, and prove performance guarantees that depend on how well the coarse space captures the user's preferences. We demonstrate substantial improvement over conventional bandit algorithms through extensive simulation as well as a live user study in the setting of personalized news recommendation.

## 1. Introduction

User feedback (e.g., ratings and clicks) has become a crucial source of training data for optimizing recommender systems. When making recommendations, one must balance the needs for exploration (gathering informative feedback) and exploitation (maximizing estimated user utility). A common formalization of such a problem is the linear stochastic bandit problem (Li et al., 2010), which models user utility as a linear function of user and content features.

Unfortunately, conventional bandit algorithms can converge slowly with even moderately large feature spaces. For instance, the well-studied LinUCB algorithm (Dani et al., 2008; Abbasi-Yadkori et al., 2011) achieves a regret bound that is linear in the dimensionality of the feature space, which cannot be improved without further assumptions.[1]

Intuitively, any bandit algorithm make recommendations that cover the entire feature space in order to guarantee learning a reliable user model. Therefore, a common approach to dealing with slow convergence is dimensionality reduction based on prior knowledge, such as previously learned user profiles, by representing new users as linear combinations of "stereotypical users" (Li et al., 2010; Yue & Guestrin, 2011).

However, if a user deviates from stereotypical users, then a reduced space may not be expressive enough to adequately learn her preferences. The challenge lies in appropriately leveraging prior knowledge to reduce the cost of exploration for new users, while maintaining the representational power of the full feature space.

Our solution is a coarse-to-fine hierarchical approach for encoding prior knowledge. Intuitively, a coarse, low-rank subspace of the full feature space may be sufficient to accurately learn a stereotypical user's preferences. At the same time, this coarse-to-fine *feature hierarchy* allows exploration in the full space when a user is not perfectly modeled by the coarse space.

We propose an algorithm, CoFineUCB, that automatically balances exploration within the coarse-to-fine feature hierarchy. We prove regret bounds that depend on how well the user's preferences project onto the coarse subspace. We also present a simple and general method for constructing feature hierarchies using prior knowledge. We perform empirical valida-tion through simulation as well as a live user study in personalized news recommendation, demonstrating that CoFineUCB can substantially outperform conventional methods utilizing only a single feature space.

[1] The regret bound is information-theoretically optimal up to log factors (Dani et al., 2008).

## 2. The Learning Problem

We study the linear stochastic bandit problem (Abbasi-Yadkori et al., 2011), which formalizes a recommendation system as a *bandit algorithm* that iteratively performs *actions* and learns from *rewards* received per action. At each iteration $t = 1, \dots, T$, our algorithm interacts with the user as follows:

- The system recommends an item (i.e., performs an action) associated with feature vector $x_t \in \mathcal{X}_t \subset \Re^D$, which encodes content and user features.
- The user provides feedback (i.e., reward) $\hat{y}_t$.

Rewards $\hat{y}_t$ are modeled as a linear function of actions $x \in \Re^D$ such that $E[\hat{y}_t|x] = w^{*\top}x$, where the weight vector $w^*$ denotes the user's (unknown) preferences. We assume feedback to be independently sampled and bounded within $[0, 1]$,[2] and that $\|x\| \leq 1$ holds for all $x$. We quantify performance using the notion of regret which compares the expected rewards of the selected actions versus the optimal expected rewards:

$$R_T(w^*) = \sum_{t=1}^{T} w^{*\top}x_t^* - w^{*\top}x_t, \quad (1)$$

where $x_t^* = \text{argmax}_{x \in \mathcal{X}_t} w^{*T}x$.[3]

We further suppose that user preferences are distributed according to some distribution $\mathcal{W}$. We can then define the expected regret over $\mathcal{W}$ as

$$R_T(\mathcal{W}) = E_{w^* \sim \mathcal{W}}\left[R_T(w^*)\right], \quad (2)$$

and the goal now for the bandit algorithm is to perform well with respect to $\mathcal{W}$. We will present an approach for optimizing (2) given a collection of existing user profiles sampled i.i.d. from $\mathcal{W}$.

## 3. Feature Hierarchies

To learn a reliable user model (i.e., a reliable estimate of $w^*$) from user feedback, bandit algorithms must make recommendations that explore the entire $D$-dimensional feature space. Conventional bandit algorithms such as LinUCB place uniform a priori importance on each dimension, which can be inefficient in practice, especially if additional structure can be assumed. We now motivate and formalize one such structure: the feature hierarchy.

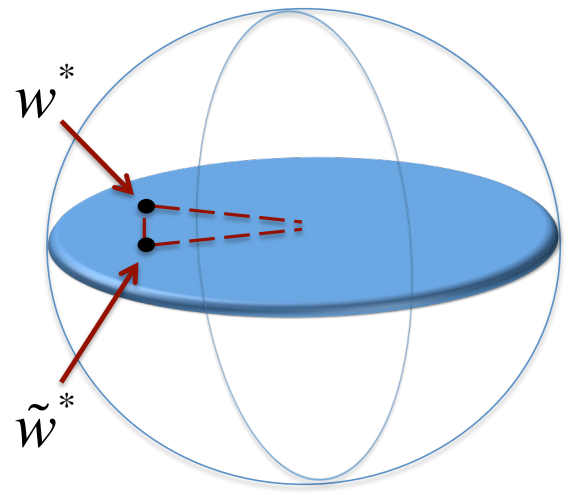

*Figure 1.* A visualization of a feature hierarchy, where $w^*$ denotes the user profile, and $\tilde{w}^*$ the projected user profile.

For example, suppose that two of the $D$ features correspond to interest in articles about baseball and cricket. Suppose also that our prior knowledge suggests that users are typically interested in one or the other, but rarely both. Then we can design a feature subspace where baseball and cricket topics project along opposite directions in a single dimension. A bandit algorithm leveraging this structure should, ideally, first explore at a coarse level to determine whether the user is more interested in articles about baseball or cricket.

We can formalize the different levels of exploration as a hierarchy that is composed of the full feature space and a subspace. We define a $K$-dimensional subspace using a matrix $U \in \Re^{D \times K}$, and denote the projection of action $x \in \Re^D$ into the subspace as

$$\tilde{x} \equiv U^\top x.$$

Likewise, we can write the user's preferences $w^*$ as

$$w^* = U\tilde{w}^* + w_\perp^*, \quad (3)$$

where we call $w_\perp^*$ the residual, or orthogonal component, of $w^*$ w.r.t. $U$. Then,

$$w^{*\top}x = \tilde{w}^{*\top}\tilde{x} + w_\perp^{*\top}x.$$

Figure 1 illustrates a feature hierarchy with a two dimensional subspace. Here, $w^*$ projects well to the subspace, so we expect $w^{*\top}x \approx \tilde{w}^{*\top}\tilde{x}$ (i.e., $\|w_\perp^*\|$ is small). In such cases, a bandit algorithm can focus exploration on the subspace to achieve faster convergence.

### 3.1. Extension to Deeper Hierarchies

For the $\ell$-th level, we define the projected $w_\ell^*$ as

$$w_{\ell-1}^* = U_\ell w_\ell^* + w_{\ell,\perp}^*.$$

Then,

$$w^* = U_1(U_2(\dots(U_L w_L^* + w_{L-1,\perp}^*)\dots w_{1,\perp}^*) + w_\perp^*.$$

[2] Our results also hold when each $\hat{y}_t$ is independent with sub-Gaussian noise and mean $w^{*\top}x_t$ (see Appendix A).

[3] Since the rewards are sampled independently, any guarantee on (1) translates into a high probability guarantee on the regret of the observed feedback $\sum_{t=1}^{T} w^{*\top}x_t^* - \hat{y}_t$.

**Algorithm 1** CoFineUCB

1: **input**: $\lambda$, $\tilde{\lambda}$, $U$, $c_t(\cdot)$, $\tilde{c}_t(\cdot)$
2: **for** $t = 1, \ldots, T$ **do**
3: Define $X_t \equiv [x_1, x_2, \ldots, x_{t-1}]$
4: Define $\tilde{X}_t \equiv U^\top X_t$
5: Define $Y_t \equiv [\hat{y}_1, \hat{y}_2, \ldots, \hat{y}_{t-1}]$
6: $\tilde{M}_t \leftarrow \tilde{\lambda} I_K + \tilde{X}_t \tilde{X}_t^\top$
7: $\tilde{w}_t \leftarrow \tilde{M}_t^{-1} \tilde{X}_t Y_t^\top$ *//least squares on coarse level*
8: $M_t \leftarrow \lambda I_D + X_t X_t^\top$
9: $w_t \leftarrow M_t^{-1}(X_t Y_t^\top + \lambda U \tilde{w}_t)$ *//least sq on fine level*
10: Define $\mu_t(x) \equiv w_t^\top x$
11: $x_t \leftarrow \operatorname{argmax}_{x \in \mathcal{X}_t} \mu_t(x) + c_t(x) + \tilde{c}_t(x)$ *//play action with highest upper confidence bound*
12: Recommend $x_t$, observe reward $\hat{y}_t$
13: **end for**

For simplicity and practical relevance, we focus on two-level hierarchies.

## 4. Algorithm & Main Results

We now present a bandit algorithm that exploits feature hierarchies. Our algorithm, CoFineUCB, is an upper confidence bound algorithm that generalizes the well-studied LinUCB algorithm, and automatically trades off between exploring the coarse and full feature spaces. CoFineUCB is described in Algorithm 1. At each iteration $t$, CoFineUCB estimates the user's preferences in the subspace, $\tilde{w}_t$, as well as the full feature space, $w_t$. Both estimates are solved via regularized least-squares regression. First, $\tilde{w}_t$ is estimated via

$$\tilde{w}_t = \operatorname*{argmin}_{\tilde{w}} \sum_{\tau=1}^{t-1} (\tilde{w}^\top \tilde{x}_\tau - \hat{y}_\tau)^2 + \tilde{\lambda} \|\tilde{w}\|^2, \tag{4}$$

where $\tilde{x}_\tau \equiv U^\top x_\tau$ denotes the projected features of the action taken at time $\tau$. Then $w_t$ is estimated via

$$w_t = \operatorname*{argmin}_{w} \sum_{\tau=1}^{t-1} (w^\top x_\tau - \hat{y}_\tau)^2 + \lambda \|w - U\tilde{w}_t\|^2, \tag{5}$$

which regularizes $w_t$ to the projection of $\tilde{w}_t$ back into the full space. Both optimization problems have closed form solutions (Lines 7 & 9 in Algorithm 1).

CoFineUCB is an optimistic algorithm that chooses the action with the largest *potential reward* (given some target confidence). Selecting such an action requires computing confidence intervals around the mean estimate $w_t$. We maintain confidence intervals for both the full space and the subspace, denoted $c_t(\cdot)$ and $\tilde{c}_t(\cdot)$, respectively. Intuitively, a valid $1-\delta$ confidence interval should satisfy the property that

$$|x^\top (w_t - w^*)| \leq c_t(x) + \tilde{c}_t(x) \tag{6}$$

holds with probability at least $1 - \delta$.

We will show that the following definitions of $c_t(\cdot)$ and $\tilde{c}_t(\cdot)$ yield a valid $1-\delta$ confidence interval:

$$\tilde{c}_t(x) = \tilde{\alpha}_t^{(v)} \left\| U^\top M_t^{-1} x \right\|_{\tilde{M}_t^{-1}} + \tilde{\alpha}_t^{(b)} \left\| \tilde{M}_t^{-1} U^\top M_t^{-1} x \right\| \tag{7}$$

$$c_t(x) = \alpha_t^{(v)} \|x\|_{M_t^{-1}} + \alpha_t^{(b)} \left\| M_t^{-1} x \right\|, \tag{8}$$

where $\tilde{\alpha}_t^{(v)}$, $\tilde{\alpha}_t^{(b)}$, $\alpha_t^{(v)}$, and $\alpha_t^{(b)}$ are coefficients that must be set properly (Lemma 1).

Broadly speaking, there are two types of uncertainty affecting an estimate, $w_t^\top x$, of the utility of $x$: variance and bias. In our setting, variance is due to the stochasticity of user feedback $\hat{y}_t$. Bias, on the other hand, is due to regularization when estimating $\tilde{w}_t$ and $w_t$. Intuitively, as our algorithm receives more feedback, it becomes less uncertain (w.r.t. both bias and variance) of its estimates, $\tilde{w}_t$ and $w_t$. This notion of uncertainty is captured via the inverse feature covariance matrices $\tilde{M}_t$ and $M_t$ (Lines 6 & 8 in Algorithm 1). Table 1 provides an interpretation of the four sources of uncertainty described in (7) and (8).

Lemma 1 below describes how to set the coefficients such that $c_t(x) + \tilde{c}_t(x)$ is a valid $1-\delta$ confidence bound.

**Lemma 1.** *Define $\tilde{S} = \|\tilde{w}^*\|$ and $S_\perp = \|w_\perp^*\|$, and let*

$$\alpha_t^{(v)} = \sqrt{\log\left(\det(M_t)^{1/2} \det(\lambda I_D)^{1/2} / \delta\right)}$$

$$\tilde{\alpha}_t^{(v)} = \lambda \sqrt{\log\left(\det\left(\tilde{M}_t\right)^{1/2} \det\left(\tilde{\lambda} I_K\right)^{1/2} / \delta\right)}$$

$$\alpha_t^{(b)} = \sqrt{2} \lambda S_\perp$$

$$\tilde{\alpha}_t^{(b)} = \lambda \tilde{\lambda} \tilde{S}.$$

*Then (6) is a valid $1-\delta$ confidence interval.*

With the confidence intervals defined, we are now ready to present our main result on the regret bound.

**Theorem 1.** *Define $\tilde{c}_t(\cdot)$ and $c_t(\cdot)$ as in (7), (8) and Lemma 1. For $\lambda \geq \max_x \|x\|^2$ and $\tilde{\lambda} \geq \max_x \|\tilde{x}\|^2$, with probability $1-\delta$, CoFineUCB achieves regret*

$$R_T(w^*) \leq \left(\beta_T \sqrt{D} + \tilde{\beta}_T \sqrt{K}\right) \sqrt{2T \log(1+T)},$$

*where*

$$\beta_T = \sqrt{D \log((1 + T/\lambda)/\delta)} + \sqrt{2\lambda} S_\perp \tag{9}$$

$$\tilde{\beta}_T = \sqrt{K \log((1 + T/\tilde{\lambda})/\delta)} + \sqrt{\tilde{\lambda}} \tilde{S}. \tag{10}$$

Lemma 1 and Theorem 1 are proved in Appendix A.

Theorem 1 essentially bounds the regret as

$$R_T(w^*) = \mathcal{O}\left(\left(\sqrt{\tilde{\lambda}} \|\tilde{w}^*\| K + \sqrt{2\lambda} \|w_\perp^*\| D\right) \sqrt{T}\right), \tag{11}$$

| Term | Interpretation |
|---|---|
| $\alpha_t^{(v)} \|x\|_{M_t^{-1}}$ | feedback variance in full space |
| $\tilde{\alpha}_t^{(v)} \left\|U^\top M_t^{-1} x\right\|_{\tilde{M}_t^{-1}}$ | feedback variance in coarse space |
| $\alpha_t^{(b)} \left\|M_t^{-1} x\right\|$ | regularization bias in full space |
| $\tilde{\alpha}_t^{(b)} \left\|\tilde{M}_t^{-1} U^\top M_t^{-1} x\right\|$ | regularization bias in coarse space |

*Table 1.* Interpretating sources of uncertainty in (7), (8).

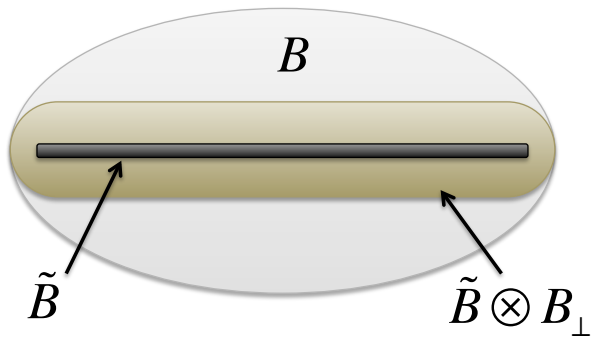

*Figure 2.* An example of confidence regions utilized by CoFineUCB and LinUCB. $B$ denotes the ellipsoid confidence region used by LinUCB. CoFineUCB maintains two ellipsoid confidence regions, $\tilde{B}$ and $B_\perp$, for subspace and full space, respectively. The joint confidence region of CoFineUCB is essentially the convolution of $\tilde{B}$ and $B_\perp$, $\tilde{B} \otimes B_\perp$, which can be much smaller than $B$.

ignoring log factors. In contrast, the conventional LinUCB algorithm only explores in the full feature space and achieves an analogous regret bound of

$$R_T(w^*) = \mathcal{O}\left(\sqrt{\lambda}\|w^*\|D\sqrt{T}\right). \tag{12}$$

Comparing (11) with (12) suggests that, when $K << D$ and $\|w_\perp^*\|$ is small, CoFineUCB suffers much less regret due to more efficient exploration. Depending on $U$, $\|\tilde{w}^*\|$ can also be much smaller than $\|w^*\|$. Section 5 describes an approach for computing such a $U$.

Intuitively, CoFineUCB enjoys a superior regret bound to LinUCB due to its use of tighter confidence regions. Figure 2 depicts a comparative example. LinUCB employs ellipsoid confidence regions. CoFineUCB utilizes confidence regions that are essentially the convolution of two smaller ellipsoids, which can be much smaller than the confidence regions of LinUCB.

## 5. Constructing Feature Hierarchies

We now show how to construct a subspace $U$ using pre-existing user profiles $W = \{w_i^*\}_{i=1}^N$, where each profile is sampled independently from a common distribution $w_i^* \sim \mathcal{W}$. In this setting, a reasonable objective is to find a $U$ that minimizes an empirical estimate of the bound on $R_T(\mathcal{W})$, which comprises $\|\tilde{w}\|$ and $\|w_\perp\|$.

Our approach is outlined in Algorithm 2. We assume that finding a $K$-dimensional subspace with low residual norms $\|w_\perp^*\|$ is straightfoward. In our experiments, we simply use the top $K$ singular vectors of $W$.

**Algorithm 2** LearnU: learning projection matrix

**input**: $W \in \Re^{D \times N}$, $K \in \{1, \ldots, D\}$
$(A, \Sigma, B) \leftarrow SVD(W)$
$U_0 \leftarrow A_{1:K}$ *//top K singular vectors*
Solve for $\Omega$ via (16) using $U_0$ and $W$
**return**: $U_0 \Omega^{1/2}$

Given an orthonormal basis $U_0 \in \Re^{K \times D}$, one can choose $U \in \texttt{span}(U_0)$ to minimize its total contribution to the regret bound in (11) over the users in $W$:

$$\underset{U \in \texttt{span}(U_0)}{\operatorname{argmin}} \ \tilde{C} \sum_{w \in W} \|\tilde{w}\|, \tag{13}$$

where $\tilde{w} \equiv (U^\top U)^{-1} U^\top w$, and $\tilde{C} = \max_x \|U^\top x\|$ constrains the magnitude of $U$.

It is difficult to optimize (13) directly, so we approximate it using a smooth formulation,[4]

$$\underset{U \in \texttt{span}(U_0): \|U\|_{Fro}^2 = K}{\operatorname{argmin}} \sum_{w \in W} \|\tilde{w}\|^2, \tag{14}$$

where we now constrain $U$ via $\|U\|_{Fro}^2 = K$.

We further restrict $U$ to be $U \equiv U_0 \Omega^{1/2}$ for $\Omega \succeq 0$. Under this restriction, (14) is equivalent to

$$\underset{\Omega: \texttt{trace}(\Omega) = K}{\operatorname{argmin}} \sum_{w \in W} \|\tilde{w}_0^\top \Omega^{-1} \tilde{w}_0\|^2, \tag{15}$$

where $\tilde{w}_0 \equiv (U_0^\top U_0)^{-1} U_0^\top w = U_0^\top w$. This formulation is akin to multi-task structure learning, where $W_0$ would denote the various tasks and $\Omega$ denotes feature relationships common across tasks (Argyriou et al., 2007; Zhang & Yeung, 2010). One can show that (15) is convex and is minimized by

$$\Omega = \frac{K}{\texttt{trace}\sqrt{\tilde{W}_0 \tilde{W}_0^\top}} \sqrt{\tilde{W}_0 \tilde{W}_0^\top}, \tag{16}$$

where $\tilde{W}_0 \equiv (U_0^\top U_0)^{-1} U_0^\top W = U_0^\top W$. See Appendix B for a more detailed derivation.

## 6. Experiments

We evaluate CoFineUCB via both simulations and a live user study in the personalized news recommendation domain. We first describe alternative methods, or baselines, for leveraging prior knowledge (pre-existing profiles $W \in \Re^{D \times N}$) that do not use a feature hierarchy. These baselines can conceptually be phrased as special cases of CoFineUCB. The key idea is to alter the feature space such that $\|w^*\|$ in the new space is small. Thus, running LinUCB in the altered feature space yields an improved bound on the regret (12), which is linear in $\|w^*\|$.

[4]One can also regularize by inserting an axis-aligned "ridge" into $W$ (i.e., $W \leftarrow [W, I_D]$).

### 6.1. Baseline Approaches

**Mean-Regularized** One simple approach is to regularize to $\bar{w}$ (e.g., the mean of $W$) when estimating $w_t$ in LinUCB. The estimation problem can be written as

$$w_t = \operatorname*{argmin}_w \sum_{\tau=1}^{t-1} (w^\top x_\tau - \hat{y}_\tau)^2 + \lambda \|w - \bar{w}\|^2. \quad (17)$$

Typically, $\|w^* - \bar{w}\| < \|w^*\|$, implying lower regret.

**Reshape** Another approach is to use LinUCB with a feature space "reshaped" via a transform $U_D \in \Re^{D \times D}$:

$$w_t = \operatorname*{argmin}_w \sum_{\tau=1}^{t-1} (w^\top U_D^\top x_\tau - \hat{y}_\tau)^2 + \lambda \|w\|^2. \quad (18)$$

As in the mean-regularization approach above, here we would like the representation of $w^*$ in the reshaped space to have a small norm. In our experiments, we use $U_D = \text{LearnU}(W, D)$ (Algorithm 2).

We can incorporate such reshaping into CoFineUCB. We first project $W$ into the space defined by $U_D$, denoted by $\hat{W}$,[5] then compute $U$ via $\text{LearnU}(\hat{W}, K)$. During model estimation, we replace (5) with

$$w_t = \operatorname*{argmin}_w \sum_{\tau=1}^{t-1} (w^\top U_D^\top x_\tau - \hat{y}_\tau)^2 + \lambda \|w - U\tilde{w}_t\|^2.$$

Incorporating reshaping into CoFineUCB can lead to a decrease in $S_\perp = \|\hat{w}_\perp^*\|$. We found the modification to be quite effective in practice; all our experiments in the following sections employ this variant of CoFineUCB.

**SubspaceUCB** Finally, we can simply ignore the full space and only apply LinUCB in the subspace. While the method seems to perform well given a good subspace (as seen in (Li et al., 2010; Chapelle & Li, 2011; Yue & Guestrin, 2011), among others), it can yield linear regret if the residual of the user's preference is strong, as we will see in the experiments.

### 6.2. Experimental Setting

We employ the submodular bandit extension of linear stochastic bandits (Yue & Guestrin, 2011) to model the news recommendation setting. Here, the algorithm must choose a set of $L$ actions and receives rewards based on both the quality as well as diversity of the actions chosen ($L = 1$ is the conventional bandit setting). Using this structured action space leads to a more realistic setting for content recommendation, since recommender systems often must recommend multiple items at a time. It is straightforward to extend CoFineUCB to the submodular bandit setting (see Appendix C).

[5] $\hat{W} \equiv \left(U_D^\top U_D\right)^{-1} U_D^\top W$.

### 6.3. Simulations

We performed simulation evaluations using data collected from a previous user study in personalized news recommendation by (Yue & Guestrin, 2011). The data includes featurized articles ($D = 100$) and $N = 77$ user profiles. We employed leave-one-out validation: for each user, the transformations $U_D$ and $U$ ($K = 5$) were trained using the remaining users' profiles. For each user, we ran 25 simulations ($T = 10000$). All algorithms used the same $U$ and $U_D$ projections, where applicable. We also compared with a variant of CoFineUCB, CoFineUCB-focus, which scales down exploration in the full space $c_t$ by a factor of 0.25.

Figure 3(a) shows the cumulative regret of each algorithm averaged over all users when recommending one article per iteration ($L = 1$). All algorithms dramatically outperform Naive LinUCB, with the exception of Mean-Regularized which performs almost identically. While Reshape shows good eventual convergence behavior, it incurs higher initial regret than the CoFineUCB algorithms and SubspaceUCB. The trends also hold when recommending multiple articles per iteration ($L = 5$), as seen in Figure 3(b).

The performance of the two variants of CoFineUCB and SubspaceUCB demonstrate the benefit of exploring in the subspace. However, Figure 3(c) reveals the critical shortfall of SubspaceUCB by comparing average cumulative regret for the ten users with the largest residual $\|w_\perp^*\|$. For these atypical users, the subspace is not sufficient to adequately learn their preferences, resulting in linear regret for SubspaceUCB.

Figure 3(d) shows the behavior of CoFineUCB as we vary $K$. Larger subspaces require more exploration, which in general leads to increased regret.

Figure 3(e) shows the behavior of CoFineUCB as we vary the scaling of exploration in the full space $c_t$ (CoFineUCB-focus is the special case where the scaling factor is 0.25). More conservative exploration in the full space tends to reduce regret. However, no exploration of the full space can lead to higher regret.

**Synthetic Dataset**. We used a 25-dimensional synthetic dataset to study the effect of mismatch between

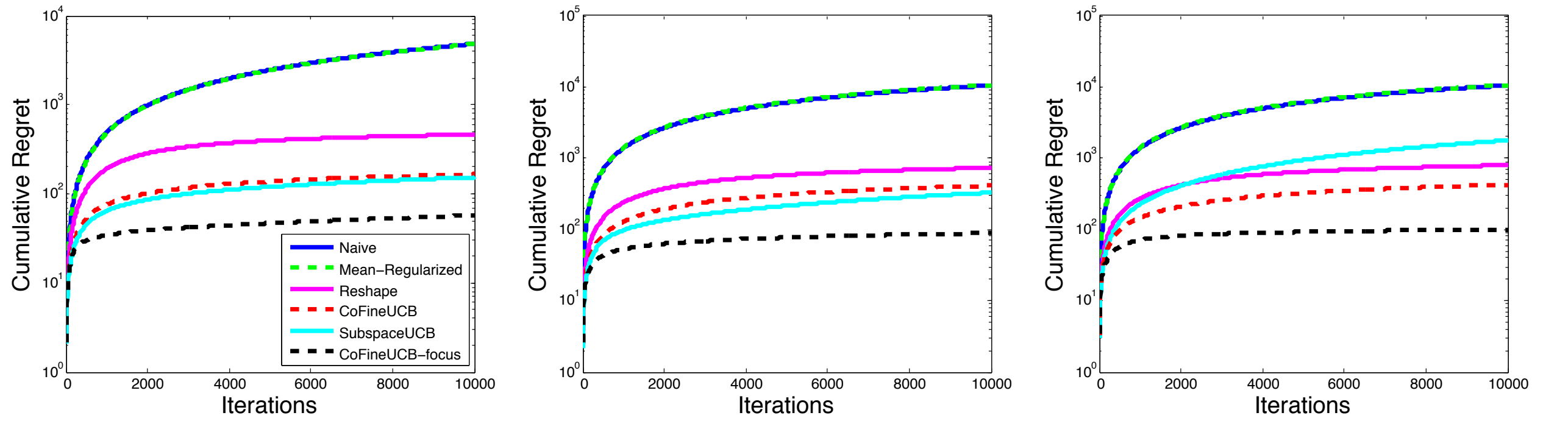

(a) All users simulation ($L = 1$) (b) All users simulation ($L = 5$) (c) Atypical users simulation ($L = 5$)

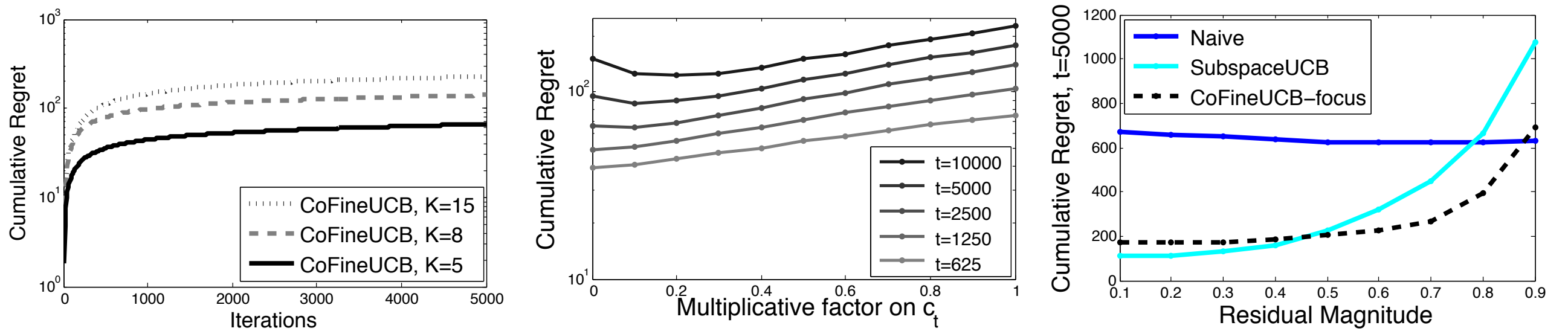

(d) CoFineUCB over varying $K$ (e) CoFineUCB over varying $c_t$ (f) Subspace mismatch versus regret

*Figure 3.* (a)–(e) Cumulative regret results for news recommendation simulation. (f) Comparison over preference vectors with varying projection residuals using synthetic simulation.

$w^*$ and $U$. This dataset allows for a more systematic analysis by forcing every $x$ and $w^*$ to have unit norm. For residual magnitude $\beta \in [0, 1]$, we sampled $w^*$ uniformly in a 5-dimensional subspace with magnitude $\sqrt{1 - \beta^2}$, and uniformly in the remaining dimensions with magnitude $\beta$. Figure 3(f) shows the regret of both SubspaceUCB and CoFineUCB-focus increase with the residual, with SubspaceUCB exhibiting more dramatic increase, beyond that of even Naive LinUCB.

### 6.4. User Study

Our user study design follows the study conducted in (Yue & Guestrin, 2011). We presented each user with ten articles per day over ten days from January 21, 2012 to February 8, 2012. Each day comprised approximately ten thousand articles. We represented articles using $D = 100$ features corresponding to topics learned via latent Dirichlet Allocation (Blei et al., 2003). For each day, articles shown are selected using an interleaving of two bandit algorithms. The user is instructed to briefly skim each article and mark each article as "interested in reading in detail" or "not interested".

We conducted the user study in two phases. Prior to the first phase, we conducted a preliminary study to collect preferences for constructing $U$ ($K = 5$). In the first phase, we compared CoFineUCB with Naive. Afterwards, we took all the user profiles learned so far to estimate a reshaping of the full space $U_D$, and compared against Reshape. Due to the short duration of each session ($T = 10$), we did not expect a meaningful comparison between CoFineUCB and SubspaceUCB, so we omitted it (We expect both methods to perform equally well in early iterations, as seen in the simulation experiments.). For each user session, we counted the total number of liked articles recommended by each algorithm. An algorithm wins a session if the user liked more articles recommended by it.

| Comparison | #Users | Win/Tie/Lose | Gain/Day |
|---|---|---|---|
| CoFineUCB v. Naive | 27 | 24 / 1 / 3 | 0.69 |
| CoFineUCB v. Reshape | 30 | 21 / 3 / 6 | 0.27 |

*Table 2.* User study comparing CoFineUCB with two baselines. All results satisfy 95% statistical confidence.

Table 2 shows that over the two stages, about 80% of the users prefer CoFineUCB. We see a smaller gain against Reshape, a stronger baseline. On average, users liked over half an additional article per day from CoFineUCB over Naive, and about a quarter addi-tional per day over Reshape. These results show that CoFineUCB is effective in reducing the amount of exploration required.

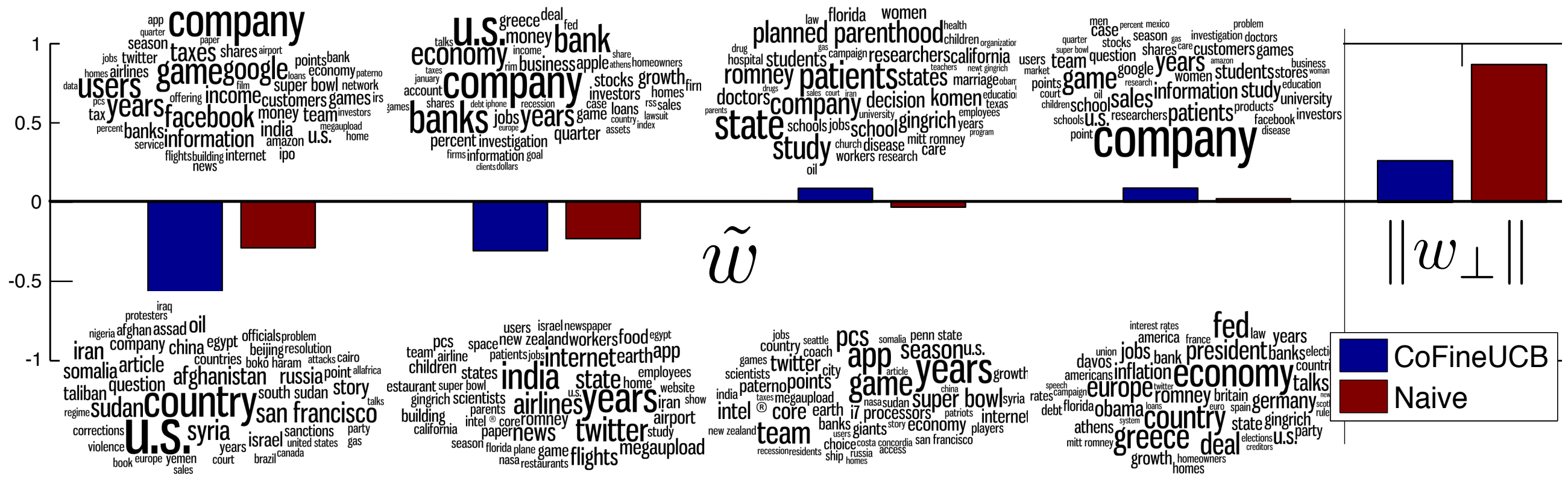

*Figure 4.* Each column of word clouds represents a dimension in the subspace. The bar lengths denote the magnitude in each dimension of preferences vectors learned by CoFineUCB (blue) and Naive LinUCB (red). The rightmost column shows the norm of residual $w_\perp^*$ of weight vectors learned by CoFineUCB and Naive LinUCB.

Figure 4 shows a representation of four dimensions of $U$ learned from user profiles. Each dimension is a combination of features, i.e., topics from LDA. In the top row, the $i$-th word cloud contains representative words from topics associated with high positive weights in $i$-th column of $U$, and the bottom row those with high negative weights. Examining Figure 4 can reveal tendencies in the user preferences collected in our study; for example, the third column shows that users interested in Republican politics also tend to follow healthcare debates, but tend to be uninterested in videogaming. Figure 4 also shows a comparison of weights estimated by CoFineUCB and Naive LinUCB for one user. Since Naive LinUCB does not utilize the subspace, the weights it estimates tend to have much higher residual norm, whereas CoFineUCB puts higher weights on the subspace dimensions.

## 7. Related Work

Optimizing recommender systems via user feedback has become increasingly popular in recent years (El-Arini et al., 2009; Li et al., 2010; 2011; Yue & Guestrin, 2011; Ahmed et al., 2012). Most prior work do not address the issue of exploration and often train with pre-collected feedback, which may lead to a biased model.

The exploration-exploitation tradeoff inherent in learning from user feedback is naturally modeled as a contextual bandit problem (Langford & Zhang, 2007; Li et al., 2010; Slivkins, 2011; Chapelle & Li, 2011; Krause & Ong, 2011). In contrast to most prior work, we focus on principled approaches for encoding prior knowledge for accelerated bandit learning.

Our work builds upon a long line of research on linear stochastic bandits (Dani et al., 2008; Rusmevichientong & Tsitsiklis, 2010; Abbasi-Yadkori et al., 2011). Although often practical, one limitation is the assumption of realizability. In other words, we assume that the true model of user behavior lies within our class.

The use of hierarchies in bandit learning is not new. For instance, the work of (Pandey et al., 2007b;a) encode prior knowledge by hierarchically clustering articles into a taxonomy. However, their setting is feature-free, which can make it difficult to generalize to new articles and users. In contrast, our approach makes use of readily available feature-based prior knowledge such as the learned preferences of existing users.

Another related line of work is that of sparse linear bandits (Abbasi-Yadkori et al., 2012; Carpentier & Munos, 2012). The assumption is that the true $w^*$ is sparse, and one can achieve regret bounds that depend on the sparsity of $w^*$. In contrast, we consider settings where user profiles are not necessarily sparse, but can be well-approximated by a low-rank subspace.

It may be possible to integrate our feature hierarchy approach with other bandit learning algorithms, such as Thompson Sampling (Chapelle & Li, 2011). Thompson Sampling is a probability matching algorithm that samples $w_t$ from the posterior distribution. Using feature hierarchies, one can define a hierarchical sampling approach that first samples $\tilde{w}_t$ in the subspace, and then samples $w_t$ around $\tilde{w}_t$ in the full space.

Our approach can be applied to many structured classes of bandit problems (e.g., (Streeter & Golovin, 2008; Cesa-Bianchi & Lugosi, 2009)), assuming that actions can be featurized and modeled linearly. For instance, our experiments demonstrated substantial improvements upon naive UCB algorithms for the linear submodular bandit problem (Yue & Guestrin, 2011).

The problem of learning a good subspace $U$ is related to finding a good regularization structure for multi-task learning (Argyriou et al., 2007; Zhang & Yeung, 2010). Given a sample of user profiles (task weights), our goal is essentially to learn a regularization structure so that future users (tasks) are solved efficiently. However, the coarse subspace of our feature hierarchy was estimated using a relatively small number of imperfectly estimated existing user profiles. A more general problem would be to learn the feature hierarchy on-the-fly as an online learning problem itself.

# 8. Conclusion

We have presented a general approach to encoding prior knowledge for accelerating contextual bandit learning. In particular, our approach employs a coarse-to-fine feature hierarchy which dramatically reduces the amount of exploration required. We evaluated our approach in the setting of personalized news recommendation, where we showed significant improvements over existing approaches for encoding prior knowledge.

**Acknowledgements**. The authors thank the anonymous reviewers for their helpful comments. The authors also thank Khalid El-Arini for help with data collection and processing. This work was supported in part by ONR (PECASE) N000141010672, ONR Young Investigator Program N00014-08-1-0752, and by the Intel Science and Technology Center for Embedded Computing.